\documentclass[10pt,twocolumn,letterpaper]{article}

\usepackage{cvpr}              

\usepackage{graphicx}
\usepackage{amsmath}
\usepackage{amssymb}
\usepackage{booktabs}
\usepackage{multirow}
\usepackage{xcolor}
\usepackage{makecell}
\usepackage{enumitem}
\usepackage[pagebackref,breaklinks,colorlinks]{hyperref}

\usepackage[capitalize]{cleveref}
\crefname{section}{Sec.}{Secs.}
\Crefname{section}{Section}{Sections}
\Crefname{table}{Table}{Tables}
\crefname{table}{Tab.}{Tabs.}

\def\cvprPaperID{0968} 
\def\confName{CVPR}
\def\confYear{2022}

\begin{document}
	
	\title{TransWeather: Transformer-based Restoration of Images Degraded by Adverse Weather Conditions   }
	
	\author{Jeya Maria Jose Valanarasu, Rajeev Yasarla, and Vishal M. Patel\\
		Johns Hopkins University\\
		
		{\tt\small \{jvalana1,ryasarl1,vpatel36\} @jhu.edu}
}
\maketitle
\vspace{-2em}
\begin{abstract}
	Removing adverse weather conditions like rain, fog, and snow from images is an important problem in many applications. Most methods proposed in the literature have been designed to deal with just removing one type of degradation. Recently, a CNN-based method  using neural architecture search (All-in-One) was proposed  to remove all the weather conditions at once. However, it has a large number of parameters as it uses multiple encoders to cater to each weather removal task and still has scope for improvement in its performance. In this work, we focus on developing an efficient solution for the all adverse weather removal problem. To this end, we propose TransWeather, a transformer-based end-to-end model with just a single encoder and a decoder that can restore an image degraded by any weather condition. Specifically, we utilize a novel transformer encoder using intra-patch transformer blocks to enhance attention inside the patches to effectively remove smaller weather degradations. We also introduce a transformer decoder with learnable weather type embeddings to adjust to the weather degradation at hand. TransWeather achieves significant improvements across multiple test datasets over both All-in-One network as well as methods fine-tuned for specific tasks. TransWeather is also validated on real world test images and  found to be more effective than previous methods. Implementation code can be found in the supplementary document. Code is available at https://github.com/jeya-maria-jose/TransWeather.         
	
\end{abstract}

\section{Introduction}
\label{sec:intro}
\setlength{\belowdisplayskip}{0pt} \setlength{\belowdisplayshortskip}{0pt}
\setlength{\abovedisplayskip}{0pt} \setlength{\abovedisplayshortskip}{0pt}
Weather conditions like rain, fog, and snow reduce the visibility and corrupt the information captured by an image. This drastically affects the performance of many computer vision algorithms like detection, segmentation and depth estimation \cite{ren2015faster, carion2020end, xie2021segformer, wang2020sdc, chen2018encoder } which are important parts of autonomous navigation and surveillance systems \cite{liang2018deep,perera2018uav,qi2018frustum,prakash2021multi}. Hence, it is essential to remove adverse weather effects
from images in order to make these vision systems more reliable. Also, a clean image without any weather degradation is desired in  photography.
\begin{figure}[!]
	\centering
	\includegraphics[width=1\linewidth, page=3]{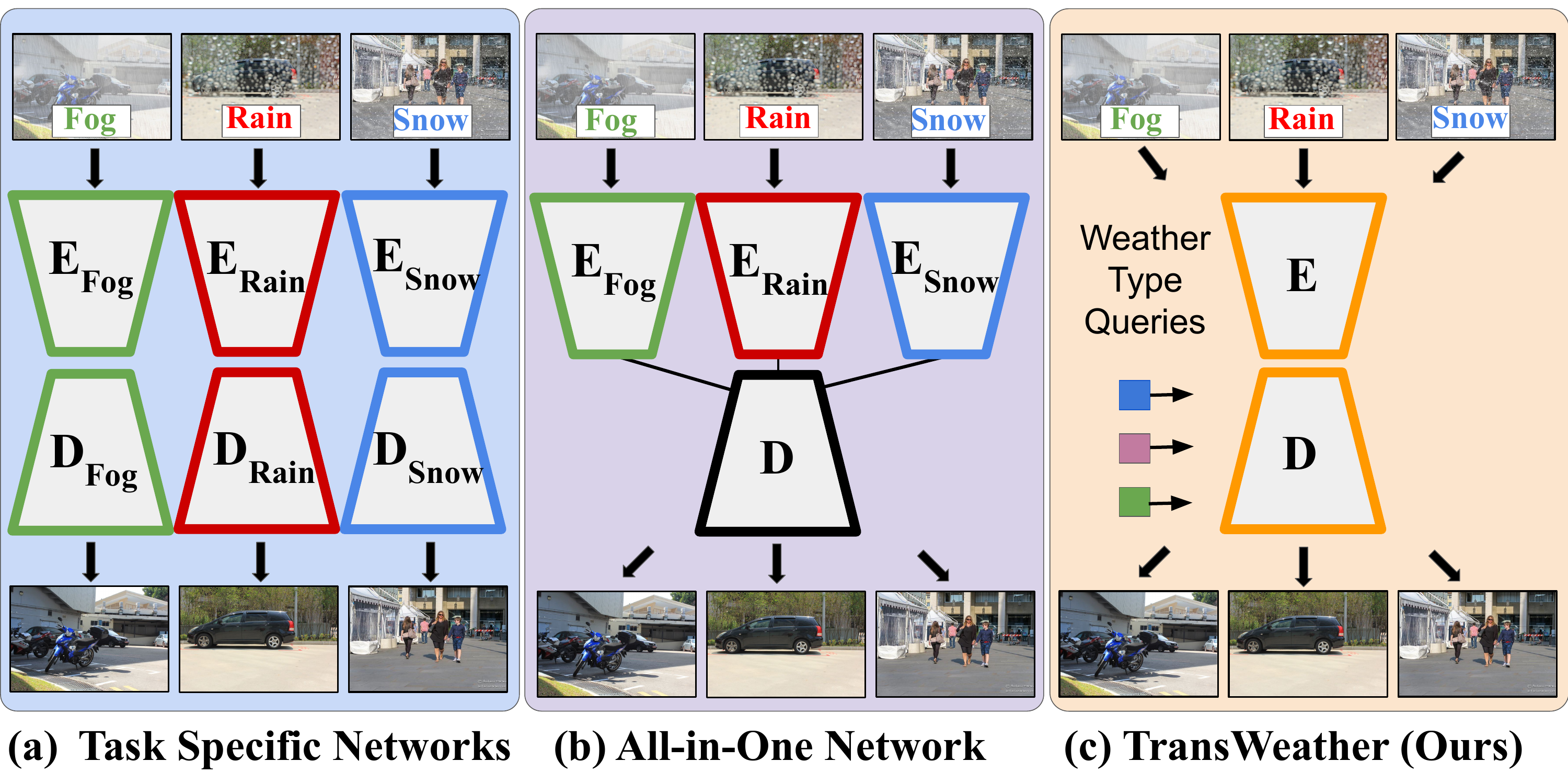}
	\caption{\textbf{Weather Removal Frameworks -} (a) Separate networks designed for each type of weather removal like rain, fog and snow. (b) All-in-One Network \cite{li2020all} proposes a framework with separate encoders for each task but a generic decoder. (c) Our proposed method, Transweather, has a single encoder and a decoder and learns weather type queries to solve all adverse weather removal efficiently. }
	\label{teaser}
\end{figure}
Early methods for weather removal involve modelling priors for weather conditions using empirical observations \cite{he2010single, roth2005fields, rudin1992nonlinear}. These priors have to be modelled separately for each weather condition and a common prior modelled for all weather conditions is not effective. Recently, Convolutional Neural Networks (CNNs) based solutions have been explored extensively for deraining \cite{zhang2018density,zhu2017joint,zhang2019image,yang2019joint,yasarla2019uncertainty,wei2019semi,wang2019spatial,qian2018attentive,fu2017removing}, dehazing \cite{ren2015faster,wu2021contrastive, dong2020multi, li2020deep, zhang2018density, zhang2019joint, zhang2021hierarchical}, desnowing \cite{liu2018desnownet, ren2017video, zhang2021deep} and raindrop removal \cite{qian2018attentive, quan2019deep, you2015adherent}. Transformer-based methods have also been explored for weather removal tasks achieving better performance than CNNs \cite{tan2021sdnet,qin2021etdnet,zhao2021hybrid}. Most of these methods just focus on one task at hand or fine-tune the model separately for each task. Although they achieve excellent performance, these are not generic solutions for all adverse weather removal problems as the networks have to trained separately for each task. This makes it difficult to adopt them for real-time systems as there have to be multiple models making it computationally complex. Also, the system would have to decide and switch between a series of weather removal algorithms (Figure \ref{teaser} (a)) making the pipeline more complicated.

Recently, Li \etal\cite{li2020all} proposed an All-in-One bad weather removal network which was the first work to propose an algorithm that takes in an image degraded by any weather condition as input and predicts the clean image. All-in-One network was tested across 3 datasets of rain, fog, and snow removal and achieved better or comparable performance than the previous methods which were tuned individually on separate datasets. All-in-One network is CNN-based and uses multiple encoders.  In particular, it uses separate encoders for the different weather degradation at hand and uses neural architecture search to find the best network to address the problem (Figure \ref{teaser} (b)). This network is still computationally complex as there are multiple encoders. To the best of our knowledge, no other methods apart from All-in-One network \cite{li2020all} have been proposed for a generic adverse weather removal in the literature. Although recent methods like MPR-Net \cite{zamir2021multi}, U-former \cite{wang2021uformer}, Swin-IR \cite{liang2021swinir} have been proposed as generic restoration networks validated on multiple datasets, they are still fine-tuned on the individual datasets and do not use a single model for all the weather removal tasks. 

In this work, we propose a single encoder-single decoder transformer network, called TransWeather, to tackle all adverse weather removal problems at once. Instead of using multiple encoders, we introduce weather type queries in the transformer decoder to learn the task (Figure \ref{teaser} (c)). Here, the multi-head self attention mechanisms take in weather type queries as input and match it with keys and values taken from features extracted from the transformer encoder. These weather type embeddings are learned along with the network to understand and adjust to the weather degradation type present in the image. The decoded features and the hierarchical features obtained form the encoder are fused and projected to the image space using a convolutional block. Thus, TransWeather just has one encoder and one decoder to learn the weather type as well as produce the clean image. Transformers are good at extracting rich global information when compared to CNNs \cite{dosovitskiy2020image}. However, we argue that when the patches are large like in ViT \cite{dosovitskiy2020image}, we fail to attend much to the information within the patch.  Weather degradations like rain streak, rain drop and snow are usually small in size and so multiple artifacts can occur within a single patch. 

To this end, we propose a novel transformer encoder with intra-patch transformer (Intra-PT) blocks. Intra-PT works on sub-patches created from the original patches and excavates features and details of smaller patches.  Intra-PT thus focuses on attention inside the main patches to remove weather degradations effectively. We use efficient self-attention mechanisms to calculate the attention between sub-patches to keep the computational complexity low. From our experiments, we find that introducing Intra-PT blocks enhances the performance of transformer and helps it adapt better to weather removal tasks. We train our network on a similar configuration as All-in-One and obtain superior performance across multiple test datasets for rain removal, snow removal, fog removal and even a combination of these weather degradations. We also outperform the methods designed specifically for these individual tasks which are finetuned on those datasets. We also show that TransWeather is fast during inference. Finally, we also test TransWeather on real-world weather degraded images, achieving excellent performance compared to the previous methods. TransWeather can act as an efficient backbone in the future for generic weather removal frameworks.

The key contributions of this work are as follows:

\begin{itemize}[topsep=0pt,noitemsep,leftmargin=*]
	\item  We propose TransWeather - an efficient solution for all adverse weather removal problem with just a single encoder and a single decoder using transformers. We propose using weather type queries to efficiently handle the All-in-One problem.
	\item  We propose a novel transformer encoder using intra-patch transformer (Intra-PT) blocks to cater to fine detail feature extraction for low-level vision tasks like weather removal. 
	\item We achieve state-of-the-art performance on multiple datasets. We also validate the effectiveness of the proposed method on real-world images. 
\end{itemize}

\section{Related Works}

Adverse weather removal problems like deraining \cite{li2018recurrent, yang2019joint, li2019heavy,kang2011automatic,zhu2017joint, yasarla2020exploring, wang2020model}, dehazing \cite{cai2016dehazenet, berman2016non, zhang2018density, fattal2014dehazing, li2018single, ren2016single}, desnowing \cite{ren2017video, liu2018desnownet, ren2017video, zhang2021deep} and rain drop removal \cite{qian2018attentive, quan2019deep, quan2021removing, you2015adherent} have been extensively explored in the literature.

\noindent \textbf{Rain Streak Removal:} Yang et al. \cite{yang2019joint} used a recurrent network to decompose rain layers to different layers of various streak types to remove the rain. Zhang et al. \cite{zhang2019image} proposed using a conditional GAN for image deraining. Yasarla et al. \cite{yasarla2020syn2real} explored using Gaussian processes to perform transfer learning from synthetic rain data to real-world rain data. Quan et al. \cite{quan2021removing} used a complementary cascaded network to remove rain streaks and raindrops in a unified framework. A more detailed survey of various rain removal methods can be found in \cite{yang2019single}.

\noindent \textbf{Fog Removal:}
Li et al. \cite{li2017aod} proposed a CNN
network considering both atmospheric light and transmission map to perform dehazing. Ren
et al. \cite{ren2018gated} proposed pre-processing a hazy image to generate multiple
inputs thus introducing color distortions to perform dehazing. Zhang and Patel \cite{zhang2018densely} proposed a pyramid CNN network for image dehazing. Zhang et al. \cite{zhang2021hierarchical} proposed a hierarchial density aware network for image dehazing.

\noindent \textbf{Rain drop Removal:} You et al. \cite{you2015adherent} proposed using temporal information to perform video-based raindrop removal. Qian et al. \cite{qian2018attentive} used an attention GAN to remove raindrop and also introduced a new dataset. Quan et al. \cite{quan2019deep} used a dual attention mechanism to remove effects of raindrops.

\noindent \textbf{Snow Removal:} Desnow-Net \cite{liu2018desnownet} was one of the first CNN-based methods proposed to remove snow from an image. Li et al. \cite{li2019stacked} proposed a stacked dense network for snow removal. Chen et al. \cite{chen2020jstasr} proposed JSTASR in which a size and transparency aware method was proposed to remove snow. Recently, DDMSNet \cite{zhang2021hierarchical} proposed a deep dense multiscale network using semantic and geometric priors for snow removal.

\noindent \textbf{All-in-One Weather Removal:} All-in-One Network \cite{li2020all} was proposed to handle multiple weather degradations using a single network. All-in-One uses a generator with multiple task-specific encoders and a common decoder. It uses a discriminator to classify the degradation type and only backpropagates the loss to specific encoders. It also uses neural architecture search to optimize the feature extraction from the  encoder.

\noindent  \textbf{Transformers in low-level vision:} Since the introduction of Vision Transformer (ViT) \cite{dosovitskiy2020image} for visual recognition, transformers have been widely adopted for various computer vision tasks \cite{zheng2021rethinking, jose2021medical, yang2021transpose, guo2021pct, liu2021video}. Especially for low-level vision, Image processing transformer \cite{chen2021pre} shows how pretraining a transformer on large-scale datasets can help in obtaining a better performance for low-level applications. U-former \cite{wang2021uformer} proposed a U-Net based transformer architecture for restoration problems. Swin-IR \cite{liang2021swinir} adopted Swin Transformer \cite{liu2021swin} for image restoration. Zhao et al. \cite{zhao2021hybrid} proposed a local-global transformer specifically for image dehazing. A multi-branch network \cite{tan2021sdnet} for deraining was also proposed based on swin transformer. In ETDNet \cite{qin2021etdnet}, an efficient transformer block to extract features in a coarse to fine way for image deraining was proposed.

Unlike the above methods, we propose a transformer-based single-encoder single-decoder network to solve all adverse weather removal tasks using a single model instance. Our Transformer encoder is also modified to cater to low-level tasks with the introduction of intra-patch transformer block. Our transformer decoder is trained with weather type queries to learn the task and uses that information to restore the clean image.

\section{Proposed Method -  TransWeather}

\begin{figure*}[]
	\centering
	\includegraphics[width=.9\linewidth, page=1]{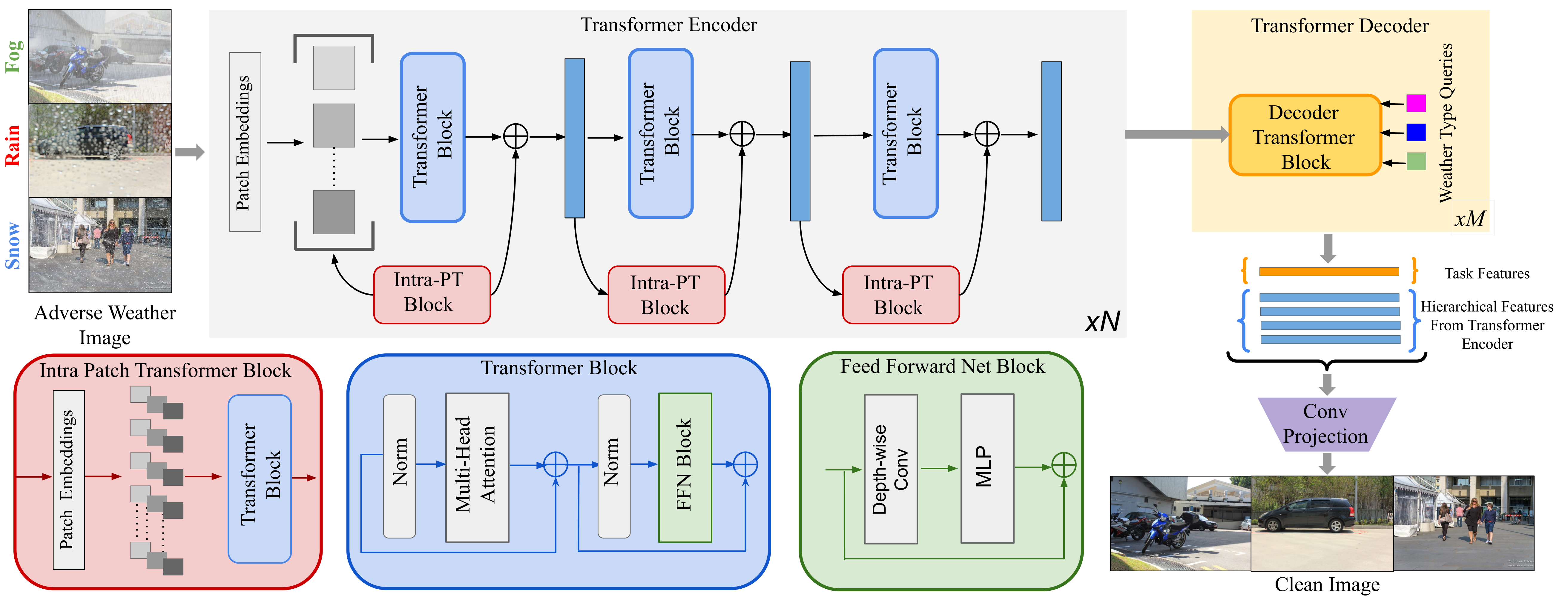}
	\vskip -10 pt
	\caption{\textbf{Overview of the proposed TransWeather network.} A degraded image is forwarded to transformer encoder to extract hierarchical features. The encoder has intra-patch transformer blocks to extract features from smaller sub-patches created from the main patch. The transformer decoder has learnable weather type queries to obtain the task feature. Then, the hierarchical features from encoder as well as the task feature from decoder are forwarded to a convolutional projection block to obtain the clean image. }  
	\label{arch}
	\vspace{-1.8em}
\end{figure*}

In the literature, different weather phenomenons have been modelled differently with regards to the underlying physics involved. Rain drop \cite{qian2018attentive} is modelled as 
\begin{equation}
	\textbf{I} = (1-\textbf{M}) \odot \textbf{B} + \textbf{R},
\end{equation}
where \textbf{I} is the degraded image, \textbf{M} is the mask, \textbf{B} is the background and \textbf{R} is the raindrop residual. Heavy rain with rain streaks and fog effect \cite{li2019heavy} is modelled as 
\begin{equation}
	\textbf{I} = \textbf{T} \odot (\textbf{B} + \sum_i^n \textbf{R}_i) + (1-\textbf{T}) \odot \textbf{A},
\end{equation}
where \textbf{T} is the transmission map produced by scattering effect,  and \textbf{A} is the atmospheric light in the scene. According to \cite{liu2018desnownet}, snow is generally modeled as 
\begin{equation}
	\textbf{I} = \textbf{M} \odot \textbf{S} + \textbf{M} \odot (1-\textbf{z}),
\end{equation}
where \textbf{z} is a mask indicating snow and \textbf{S} corresponds to snow flakes. All-in-One method \cite{li2020all} generalizes the adverse weather removal problem as
\begin{equation}
	\textbf{B} = \textbf{D}(\textbf{E}_p(\textbf{I}_p)),
\end{equation}
where \textbf{E} corresponds to the encoder and \textbf{D} corresponds to the decoder. $p$ represents the weather type present in the image. Note that for each weather type a different encoder is used. In this work, we follow a similar formulation of all adverse weather removal as
\begin{equation}
	\textbf{B} = \textbf{T}(\textbf{I}_p),
\end{equation}
where \textbf{T} corresponds to TransWeather which consists of a weather agnostic encoder and decoder network unlike All-in-One Network. The weather type queries are learnt along with the parameters of $\textbf{T()}$ thus making the problem setup more generic. We motivate this setup because a problem as generic as weather removal cannot be addressed
by merely seeking for perfection on solving individual
tasks. This formulation not only makes the process computationally efficient, but also helps in using complimentary information between the tasks to further improve the performance. Furthermore, it is also grounded with regards to how human vision works as our visual cortex can perform multiple tasks without any difficulty.  This view is widely agreed in neuro-biology as the visual cortex does not have different modules for different perception tasks \cite{mcmanus2011adaptive, li2004perceptual}.

\subsection{Network Architecture}
Given a degraded image \textbf{I} of size $H \times W \times 3$, we first divide it into patches. We then feed forward the patches to a transformer encoder containing transformer blocks at different stages. Across each stage, the resolution is reduced to make sure the transformer learns both coarse and fine information. We then use a transformer decoder block that uses the encoded features as keys and values while using learnable weather type query embeddings as queries. The extracted features are then passed through a convolutional projection block to get the clean image of dimensions $H \times W \times 3$. An overview of the network architecture of TransWeather can be found in Figure \ref{arch}. In the following sections, we describe these components in detail.

\subsubsection{Transformer Encoder}

We generate a hierarchical feature representation of the input image by extracting multi-level features in the transformer encoder. The features are extracted at different stages in the encoder thus facilitating extraction of both  high-level and low-level features. Across each stage, we perform overlapped patch merging \cite{xie2021segformer}. Using this we combine overlapping feature patches to get features of the same size as that of non-overlapped patches before passing the features to the next stage.  

\noindent \textbf{Transformer Block:} In each transformer block, we use multi-head self-attention layers and feed forward networks to calculate the self-attention features. The computation can be summarized as:
\begin{equation}
	\textbf{T}_i(\textbf{I}_i) = FFN(MSA(\textbf{I}_i)+\textbf{I}_i),
\end{equation}
where $\textbf{T}()$ represents the transformer block, $FFN()$ represents the feed forward network block, $MSA()$ represents multi head self-attention, $\textbf{I}$ is the input and $i$ represents the stage in the encoder. Similar to the original self-attention network, the heads of queries ($\textbf{Q}$), keys ($\textbf{K}$) and values ($\textbf{V}$) have same dimensions and are calculated as:
\begin{equation}
	\text{Attn}(\textbf{Q},\textbf{K},\textbf{V}) = \text{softmax}\left(\frac{\textbf{QK}^T}{\sqrt{d}}\right)\textbf{V},
\end{equation}
where \textit{d} represents the dimensionality. Note that we have multiple attention heads in each transformer block and that number is a hyper-parameter which we  vary across each stage in the transformer encoder. More details regarding the hyper-parameter settings can be found in the supplementary document.  We reduce the complexity of the original self-attention from $O(N^2)$ to $O(\frac{N^2}{R})$ by introducing a reduction ratio \textit{R} \cite{wang2021pyramid}. We reshape the keys into a dimension from a dimension of $(N,C)$ to a dimension of $(\frac{N}{R}, C.R) $. We then use a linear layer to get the second dimension back to $C$ from $C.R$. Hence, the keys get a dimension of $\frac{N}{R} \times C$ thus reducing the complexity while calculating the self attention. The self-attention features are then passed to a FFN block. 
The FFN block used here has a slight variation from ViT as we introduce using depth-wise convolution to MLP inspired from \cite{wu2021cvt,xie2021segformer, li2021localvit}. Using depth-wise convolution here helps bring locality information and provide positional information for transformers as shown in \cite{xie2021segformer}. The computation in the FFN block can be summarized as follows:
\begin{equation}
	\nonumber  FFN_i(\textbf{X}_i) = MLP(GELU(DWC(MLP(\textbf{X}_i)))) + \textbf{X}_i,    
\end{equation}
where $\textbf{X}$ refers to self-attention features, $DWC$ is depth-wise convolution \cite{chollet2017xception}, $GELU$ is Gaussian error linear units \cite{hendrycks2016gaussian}, $MLP$ is multi-layer perceptron, $i$ indicates the stage.

\noindent \textbf{Intra-Patch Transformer Block:} The intra-patch transformer blocks are present in between each stage in the transformer encoder. These blocks take in the sub-patches created from the original patches as the input. These sub-patches are fixed in dimensions half of height and width of the original patch. Intra-PT also utilizes a similar transformer block as explained above. We use a high $R$ value to make the computation very efficient in the Intra-PT block. Intra-PT block helps in extracting fine details helpful in removing smaller degradation as we operate on smaller patches. Note that the Intra-PT block creates patches at the feature level except at the first stage where it is done at the image level. The output self-attention features from the Intra-PT block are added to the self-attention features from the main block across the same stage. The formulation of feed forward process in our transformer encoder can be summarized as follows:
\begin{equation}
	\textbf{Y}_i = MT_i(\textbf{X}_i) + IntraPT_i(P(\textbf{X}_i))
\end{equation}
where $\textbf{I}$ is input to the transformer across each stage, $\textbf{Y}$ is the output across each stage,  $MT()$ is the main transformer block, $IntraPT$ is the intra-patch transformer block, $P()$ corresponds to the process of creating sub-patches from the input patches and $i$ denotes the stage. 
\vspace{-1em}
\subsubsection{Transformer Decoder}

\begin{figure}[htp!]
	\centering
	\vspace{-1.3em}
	\includegraphics[width=.75\linewidth, page=1]{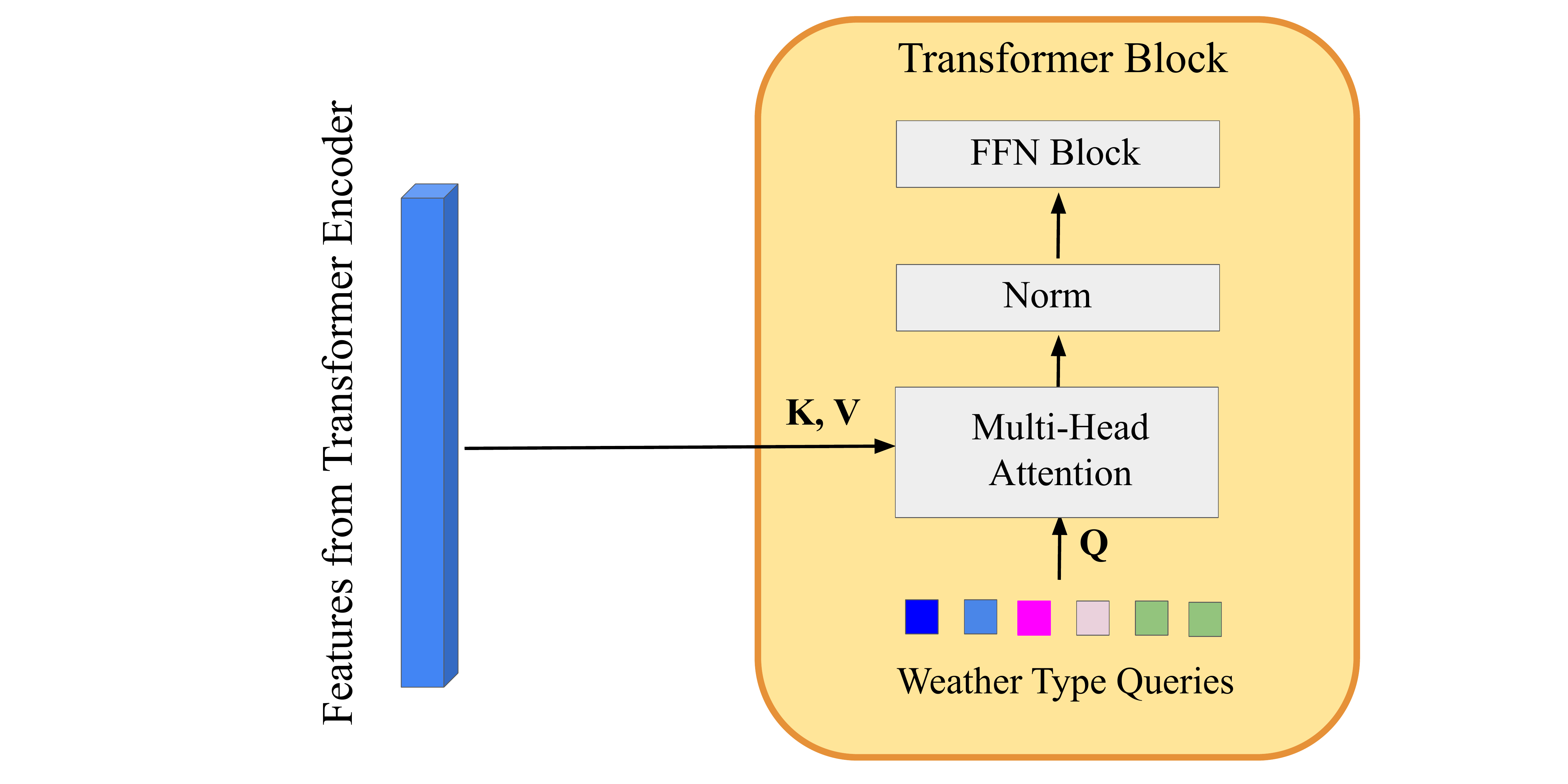}
	\vskip -12 pt
	\caption{\textbf{Configuration of the transformer block in the  decoder.} The queries here are learnable embeddings representing the weather type while the keys and values are features taken from the last stage of the transformer encoder.  }
	\label{decoder}
	\vspace{-1.6em}
\end{figure}

In the original transformer decoder \cite{vaswani2017attention}, an autoregressive decoder is used to predict the output sequence one element at a time. Detection transformer (DETR) \cite{carion2020end} uses object queries to decode the box coordinates and class labels to produce the final predictions. Inspired from them, we define weather type queries to decode the task, predict a task feature vector and use it to restore the clean image. These weather type queries are learnable embeddings which are learnt along with the other parameters of our network. These queries attend to the feature outputs from the transformer encoder. The transformer decoder here operates at a single stage but has multiple blocks.  We illustrate the transformer decoder block in Figure \ref{decoder}. These transformer blocks are similar to  encoder-decoder transformer blocks \cite{vaswani2017attention}. Unlike self-attention transformer block where $\textbf{Q}$, $\textbf{K}$ and $\textbf{V}$ are taken from the same input, here $\textbf{Q}$ is weather type learnable embedding  while $\textbf{K}$ and $\textbf{V}$ are the features taken from the last stage of the transformer encoder. The output decoded features represent the task feature vector and are fused with the features extracted across the transformer encoder at each stage. All of these features are forwarded to the convolutional tail to reconstruct the clean image.

\subsubsection{Convolutional Projection Block}
The set of hierarchical transformer encoder features and task features from the transformer decoder are passed through a set of 4 convolutional layers to output the clean image. We also use an upsampling layer before every convolutional layer to get back to the original image size. We also have skip connections across each stage in the convolutional tail from the transformer encoder. We also use a \textit{tanh} activation function in the final layer.

\subsection{Loss}

\begin{table*}[!]
	\centering
	\resizebox{1.2\columnwidth}{!}{%
		\begin{tabular}{c|c|c|c|c}
			\Xhline{4\arrayrulewidth}
			Type                                                                      & Method                    & Venue     & PSNR ($\uparrow$)          & SSIM  ($\uparrow$)          \\ \Xhline{4\arrayrulewidth}
			
			\multirow{8}{*}{\begin{tabular}[c]{@{}c@{}}Task \\ Specific\end{tabular}} & DetailsNet + Dehaze (DHF) \cite{fu2017removing} & CVPR 2017 & 13.36          & 0.5830          \\
			& DetailsNet + Dehaze (DRF) \cite{fu2017removing} & \textit{CVPR 2017} & 15.68          & 0.6400           \\
			& RESCAN + Dehaze (DHF) \cite{li2018recurrent}    & \textit{ECCV 2018} & 14.72          & 0.5870           \\
			& RESCAN + Dehaze (DHF) \cite{li2018recurrent}     & \textit{ECCV 2018} & 15.91          & 0.6150           \\
			& pix2pix \cite{isola2017image}                  & \textit{CVPR 2017} & 19.09          & 0.7100           \\
			& HRGAN \cite{li2019heavy}                    & \textit{CVPR 2019} & 21.56          & 0.8550          \\
			
			& Swin-IR \cite{liang2021swinir}                    & \textit{CVPR 2021} & 23.23          & 0.8685          \\
			& MPRNet \cite{zamir2021multi}                   & \textit{CVPR 2021} & 21.90              & 0.8456                 \\ \Xhline{3\arrayrulewidth}
			
			\multirow{2}{*}{\begin{tabular}[c]{@{}c@{}}Multi\\ Task\end{tabular}}     & All-in-One \cite{li2020all}                & \textit{CVPR 2020} & 24.71          & 0.8980           \\
			& TransWeather              & -         & 27.96 & 0.9509  \\ \Xhline{3\arrayrulewidth}
			
		\end{tabular}
	}
	
	\caption{\textbf{Quantitative Comparison on the Test1 (rain+fog) dataset based on PSNR and SSIM.} DHF represents De-Hazing First and DRF represents De-Raining First. $\uparrow$ means higher the better.}
	
	\label{test1}
\end{table*} 

Our network is trained in an end-to-end fashion using a smooth L1-loss between the prediction ($\hat{\textbf{I}}$) and the ground truth ($\textbf{G}$) defined as follows:
\begin{equation}
	\mathcal{L}_{smooth L_{1}}=\left\{\begin{array}{ll}
		0.5 \textbf{E}^{2} & \text { if }|\textbf{E}|<1 \\
		|\textbf{E}|-0.5 & \text { otherwise },
	\end{array}\right.
\end{equation}
where $\textbf{E} = \hat{\textbf{I}} - \textbf{G}$. We also add a perceptual loss that measures the discrepancy between the features of prediction and the ground truth. We extract these features using a VGG16 network \cite{simonyan2014very}  pretrained on ImageNet. We extract features from the $3^{rd}$, $8^{th}$ and $15^{th}$ layers of VGG16 to calculate the perceptual loss. The perceptual loss is formulated as  follows
\begin{equation}
	\nonumber     \mathcal{L}_{perceptual} = \mathcal{L}_{MSE}(VGG_{3,8,15}(\hat{\textbf{I}}),VGG_{3,8,15}(\textbf{G})).
\end{equation}
The total loss can be summarized as follows
\begin{equation}
	\mathcal{L}_{total} = \mathcal{L}_{smooth L_{1}} + \lambda \mathcal{L}_{perceptual},
\end{equation}
where $\lambda$ is a weight that controls the contribution of $\mathcal{L}_{perceptual}$  and L1-loss on the overall loss.


\section{Experiments}

We conduct extensive experiments to show the effectiveness of our proposed method. In what follows, we explain the datasets, implementation details, experimental settings,  results and comparison with state-of-the-art methods.

\subsection{Datasets}
We train our network on a combination of images degraded from a variety of adverse weather conditions similar to All-in-One Network \cite{li2020all}. We follow the same training set distribution used in All-in-One for fair comparison. The training data consists of 9,000 images sampled from Snow100K \cite{liu2018desnownet}, 1,069 images from Raindrop \cite{qian2018attentive} and 9,000 images of Outdoor-Rain \cite{li2019heavy}. Snow100K has synthetic images degraded by snow, raindrop has real raindrop images and Outdoor-Rain has synthetic images degraded by both fog and rain streaks. We term this combination of training data as ``All-Weather" for better representation.

We test our methods on both synthetic and real-world datasets. We use the Test1 dataset \cite{li2020all, li2019heavy}, the RainDrop test dataset \cite{qian2018attentive} and the Snow100k-L test set \cite{liu2018desnownet} for testing our method. In addition, we also evaluate on real-world images degraded by rain streaks and rain drops.

\subsection{Implementation Details}
We implement our method using Pytorch framework \cite{NEURIPS2019_9015} and train it using an NVIDIA RTX 8000 GPU. We use an Adam optimizer \cite{kingma2014adam} and a learning rate of 0.0002. We use a learning rate scheduler that anneals the learning rate by 2 after 100 and 150 epochs. The network is trained for a total of 200 epochs with a batch size of 32. Other hyper-parameters regarding the TransWeather architecture can be found in the supplementary document.

\begin{table}[]
	\centering
	\resizebox{1\columnwidth}{!}{%
		\begin{tabular}{c|c|c|c|c}
			\Xhline{4\arrayrulewidth}
			
			Type                                                                      & Method       & Venue     & PSNR ($\uparrow$)          & SSIM  ($\uparrow$)          \\ \Xhline{4\arrayrulewidth}
			
			\multirow{5}{*}{\begin{tabular}[c]{@{}c@{}}Task \\ Specific\end{tabular}} & DetailsNet \cite{fu2017removing}    & \textit{CVPR 2017} & 19.18          & 0.7495          \\
			& DesnowNet \cite{liu2018desnownet}   & TIP 2018 & 27.17          & 0.8983          \\
			& JSTASR \cite{chen2020jstasr} & ECCV 2020 & 25.32          & 0.8076          \\
			& Swin-IR \cite{liang2021swinir} & CVPR 2021 & 28.18          & 0.8800          \\
			& DDMSNET \cite{zhang2021deep}         & TIP 2021 & 28.85          & 0.8772          \\ \Xhline{3\arrayrulewidth}
			
			\multirow{2}{*}{\begin{tabular}[c]{@{}c@{}}Multi \\ Task\end{tabular}} & All-in-One \cite{li2020all}   & CVPR 2020 & 28.33          & 0.8820          \\
			& TransWeather & -         & 28.48 & 0.9308\\ \Xhline{3\arrayrulewidth}
			
		\end{tabular}
	}
	
	\caption{\textbf{Quantitative Comparison on the SnowTest100k-L test dataset based on PSNR and SSIM.} . $\uparrow$ means higher the better.} 
	
	\label{snow}
\end{table}

\subsection{Comparison with state-of-the-art methods}

First, we compare our method with state-of-the-art methods which are designed specifically for each task: rain drop removal, snow removal and rain+haze removal. For rain drop removal, we compare the performance with state-of-the-art methods like Attention GAN \cite{qian2018attentive}, Quan et al. \cite{quan2019deep}, and complementary cascaded network (CCN) \cite{quan2021removing}. For snow removal, we compare with Desnow-Net \cite{liu2018desnownet}, JSTASR \cite{chen2020jstasr} and  Deep Dense Multi-Scale Network (DDMSNet) \cite{zhang2021deep}. For rain+fog removal, we compare with HRGAN \cite{li2019heavy}, Details-Net \cite{fu2017removing}, Recurrent squeeze-and-excitation context aggregation
Net (RESCAN) \cite{li2018recurrent}, and Multi-Stage Progressive Restoration Network (MPRNet) \cite{zamir2021multi}. We also compare with a recent transformer network Swin-IR \cite{liang2021swinir} for all datasets. Note that all these methods are single-task handling networks which are fine-tuned for specific datasets.

\begin{figure*}[]
	\centering
	\includegraphics[width=0.95\linewidth, page=1]{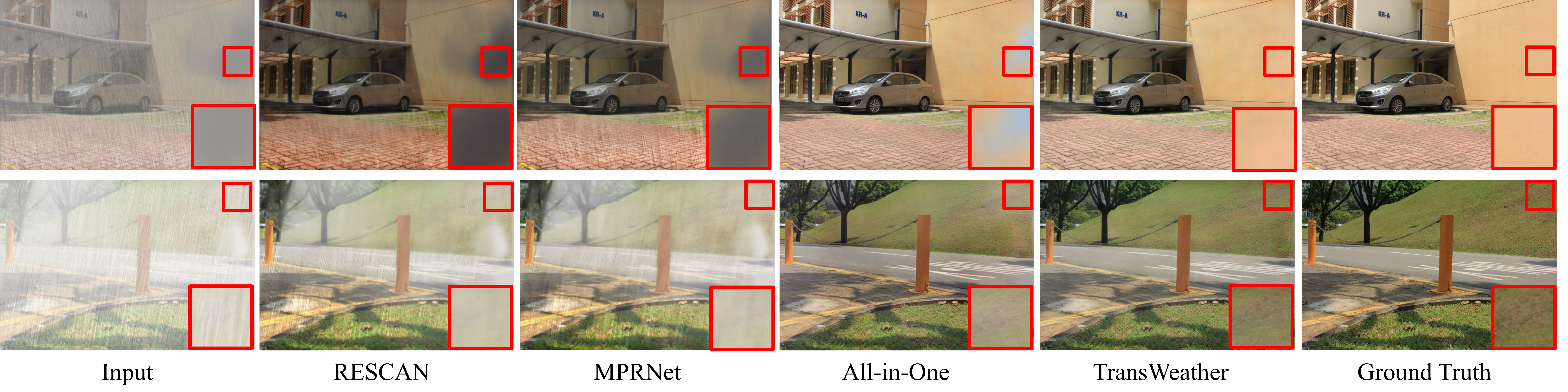}
	\vskip -12 pt
	\caption{\textbf{Sample qualitative results on the Test1 dataset.}  Red box corresponds to the zoomed-in patch for better comparison.  }
	\label{test1res}
	\vspace{-1.8em}
\end{figure*}

We also compare the performance of our method with All-in-One network \cite{li2020all} which is trained to perform all the above tasks with a \textit{single model instance}. Our method TransWeather is also trained to perform all these tasks using a \textit{single model instance}. 
\begin{table}[]
	\centering
	\resizebox{1\columnwidth}{!}{%
		
		\begin{tabular}{c|c|c|c|c}
			\Xhline{4\arrayrulewidth}
			
			Type                                                                      & Method       & Venue     & PSNR ($\uparrow$)          & SSIM ($\uparrow$)           \\ \Xhline{4\arrayrulewidth}
			
			\multirow{5}{*}{\begin{tabular}[c]{@{}c@{}}Task \\ Specific\end{tabular}} & Pix2pix \cite{isola2017image}     & \textit{CVPR 2017} & 28.02          & 0.8547          \\
			& Attn. GAN \cite{qian2018attentive}   & \textit{CVPR 2018} & 30.55          & 0.9023          \\
			& Quan et al. \cite{quan2019deep}  & \textit{ICCV 2019} & 31.44          & 0.9263          \\
			& Swin-IR \cite{liang2021swinir}  & \textit{CVPR 2021} & 30.82          & 0.9035          \\
			& CCN \cite{quan2021removing}          & \textit{CVPR 2021} & 31.34          & 0.9500          \\ \Xhline{3\arrayrulewidth}
			
			\multirow{2}{*}{\begin{tabular}[c]{@{}c@{}}Multi \\ Task\end{tabular}} & All-in-One \cite{li2020all}  & \textit{CVPR 2020} & 31.12          & 0.9268          \\
			& TransWeather & -         & 28.84 &0.9460\\ \Xhline{3\arrayrulewidth}
			
		\end{tabular}
	}
	\vskip -10 pt
	\caption{\textbf{Quantitative comparison on the RainDrop test dataset based on PSNR and SSIM.}  $\uparrow$ means higher the better.}
	
	\label{raindrop}
\end{table}
\subsubsection{Referenced Quality Metrics}
We use PSNR and SSIM to evaluate the performance of different models. We tabulate the quantitative results in terms of PSNR and SSIM in Tables 1, 2, and 3 while evaluating on the Test1 (fog+rain removal), Snow100K-L (snow removal) and RainDrop (rain drop removal) test datasets, respectively. As Test1 has both fog and rain, we sequentially apply deraining and dehazing methods for fair comparison on this dataset. For example, while using Details-Net and RESCAN for deraining, we apply Multi-scale boosted dehazing network (MSBDN) ~\cite{dong2020multi} for dehazing. Note that from our experiments we found MSBDN to be the best performing network for dehazing. We compare the performance while applying  deraining first, then dehazing and also vice-versa. We train Swin-IR and MPRNet directly on ``Outdoor-Rain" (training split of Test1) and test it on Test1 for fair comparison. Similarly, Swin-IR was trained on Snow100K dataset, RainDrop and tested on SnowTest100k-L, RainDrop test datasets respectively.  It can be noted that some recent methods like CCN and DDMSNet when fine-tuned on the individual datasets outperform All-in-One. TransWeather is observed to achieve an observed performance when compared to these methods.

\begin{figure}[]
	\centering
	\includegraphics[width=1\linewidth, page=1]{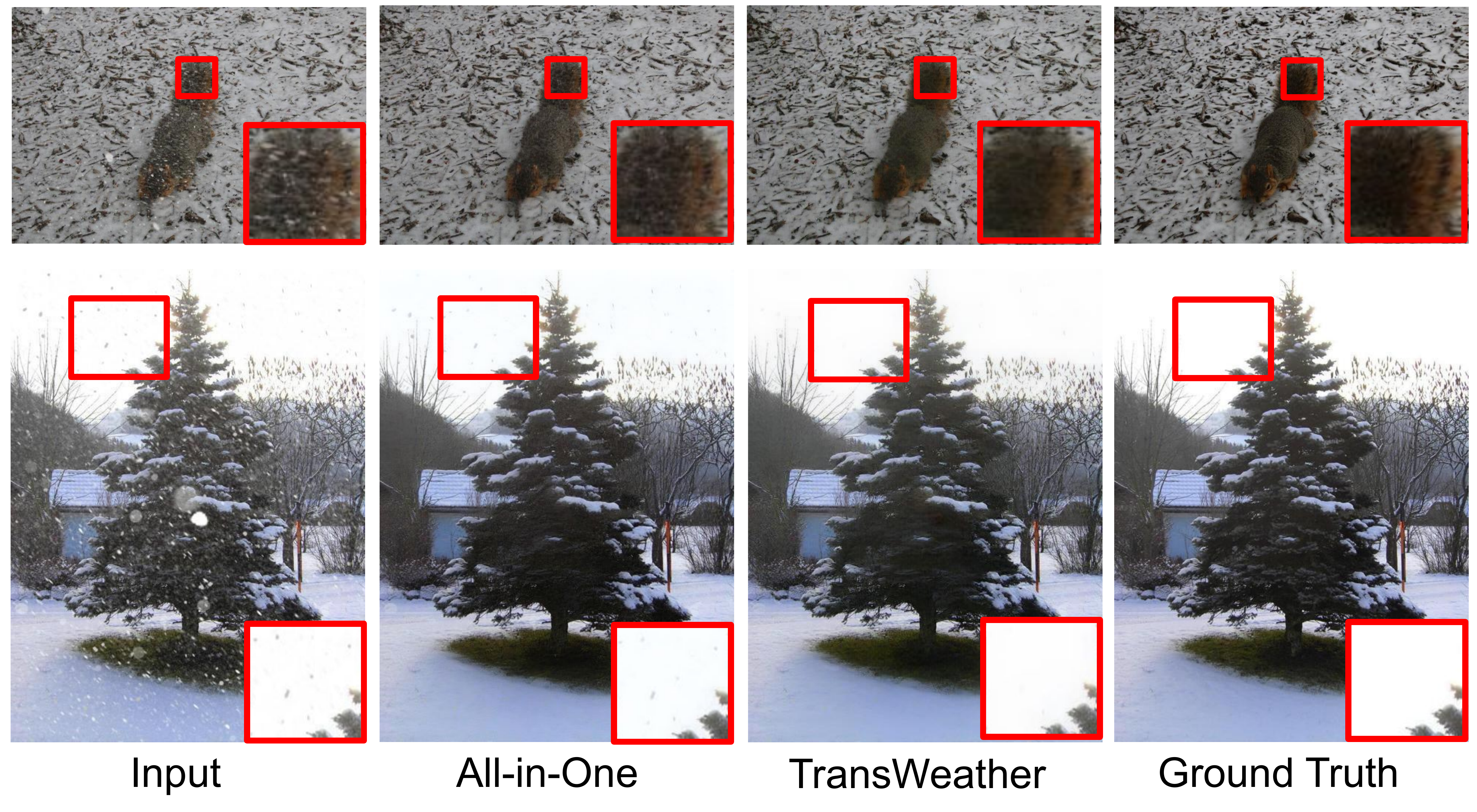} 
	
	\vskip -12 pt
	\caption{\textbf{Sample qualitative results on the Snow100k-L dataset.}  Red box corresponds to the zoomed-in patches. \label{snow} 
	}
	
\end{figure}

\begin{figure*}[]
	\centering
	\includegraphics[width=0.95\linewidth, page=1]{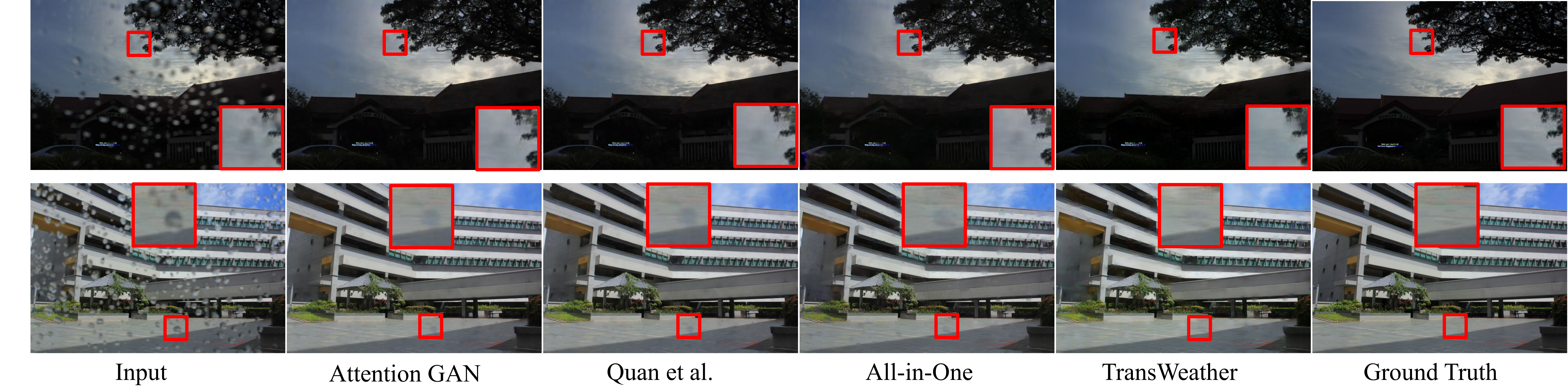} 
	
	\vskip -12 pt
	\caption{\textbf{Sample qualitative results on the RainDrop dataset.}  Red box corresponds to the zoomed-in patches for better comparison. \label{drop} 
	}
	
\end{figure*}

\subsubsection{Visual Quality Comparison}

\begin{figure*}[]
	\centering
	\includegraphics[width=0.95\linewidth, page=1]{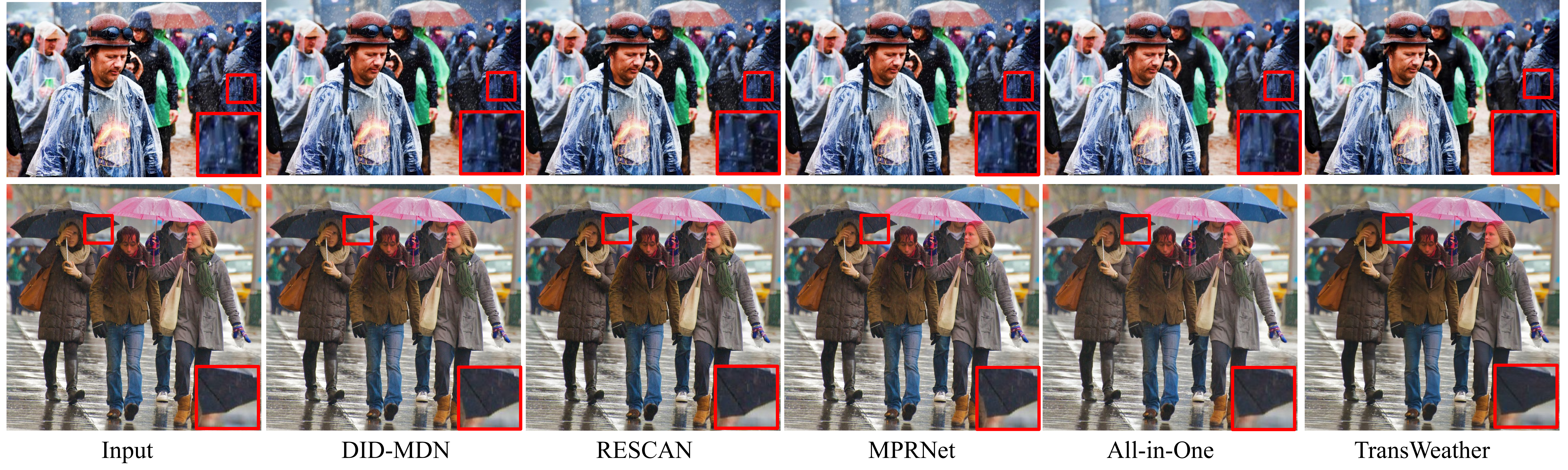} 
	
	\vskip -12 pt
	\caption{\textbf{Sample qualitative results on the Real-World images.}  Red box corresponds to the zoomed-in patch for better comparison. Note that these are real-world images with no availability of ground truth. \label{real} 
	}
	\vspace{-2em}
\end{figure*}

\noindent \textbf{Synthetic Images} We illustrate the predictions from synthetic test datasets like Test1 and Snow100k-L in Figures 4 and 5. It can be seen that Transweather achieves visually pleasing results compared to the previous methods. It works very well in removing both fog and rain streaks as can be seen in Figure  4   while other methods including All-in-One fail to remove at least one of the degradations.  It can be seen from Figure 5 that our method removes even the snow particles which are very small in structure while All-in-One has hard time removing them.

\noindent \textbf{Real-World Images} We illustrate the predictions from real test datasets like RainDrop and Real-World images in Figures 6 and 7. It can be seen in both the figures that Transweather removes even the finest rain streaks or drops when compared to the previous methods.

\section{Discussions}

\noindent \textbf{Ablation Study:} We conduct an ablation study to understand the contributions of individual components proposed in the  TransWeather architecture. We start with a base transformer encoder architecture and a conv tail. We call this configuration Transformer Base. We then convert the transformer encoder to hierarchical transformer (HE) encoder to extract both high-level and low-level features where we perform patch merging between each stage in the transformer encoder. We then add the intra patch transformer block (Intra-PT) in the encoder. Then we add learnable weather type queries and a transformer decoder block to learn the task embeddings. This configuration corresponds to the TransWeather architecture. All of these models are trained on All-Weather and tested on the Raindrop test dataset. The results of ablation study can be found in Table \ref{ablation}. It can be observed that each individual contribution of this work helps in improving the performance.

\begin{table}[htbp]
	\centering
	\resizebox{0.9\columnwidth}{!}{%
		
		\begin{tabular}{c|c|c}
			\Xhline{4\arrayrulewidth}
			
			Method            & PSNR ($\uparrow$)  & SSIM ($\uparrow$)  \\ \Xhline{4\arrayrulewidth}
			
			Transformer Base  & 30.12 & 0.8512 \\
			+ HE              & 31.62 & 0.8671 \\
			+ HE + Intra-PT             & 32.37 & 0.9463 \\
			+ HE + Intra-PT +Weather Queries & 34.55 & 0.9502 \\ \Xhline{3\arrayrulewidth}
			
		\end{tabular}
		
	}
	\vskip -9 pt
	\caption{\textbf{Ablation Study on the RainDrop test dataset.} HE denotes converting to hierarchical transformer encoder and Intra-PT represents intra-patch transformer blocks. }
	\label{ablation}
	\vspace{-1em}
\end{table}

\begin{figure}[htbp]
	\centering
	\includegraphics[width=1\linewidth, page=1]{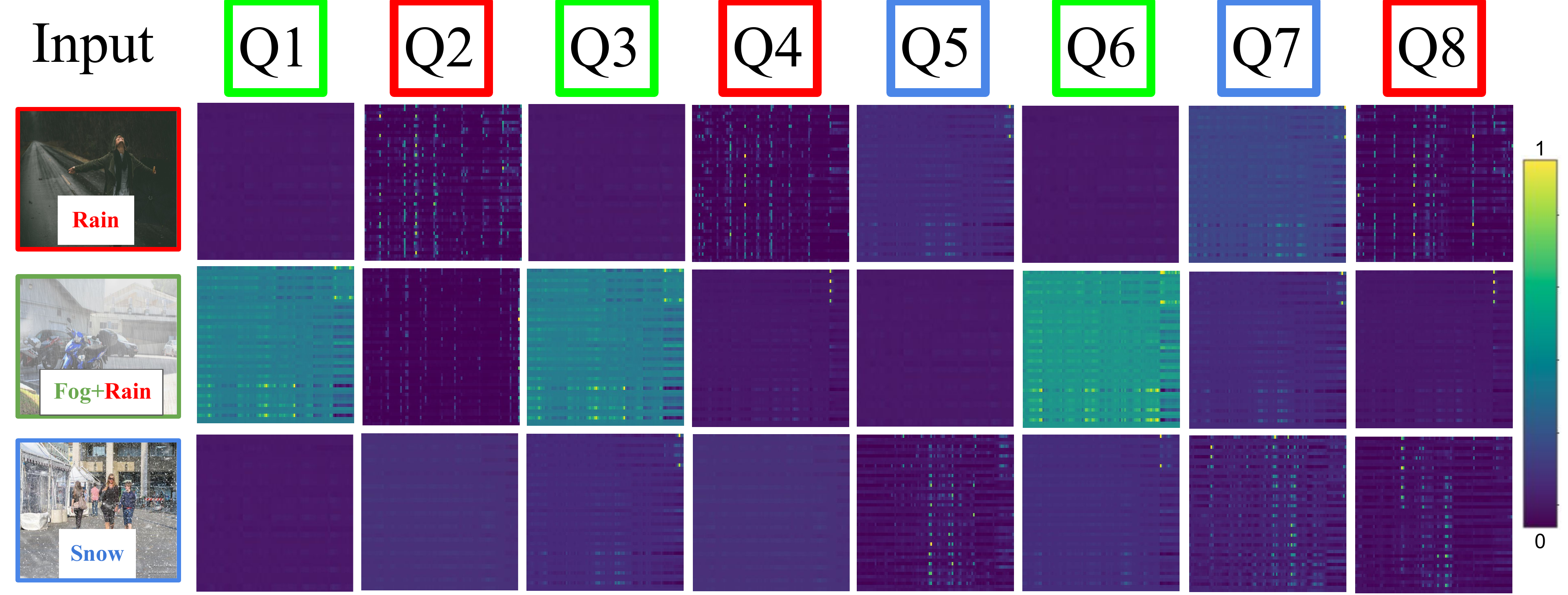} 
	\vskip -10 pt
	\caption{\textbf{Attention maps with respect to different queries.} Rows correspond to input image with different weather degradations and the columns correspond to attention maps with different queries (\textit{Q1} to \textit{Q8}). \textbf{\textcolor{red}{Red}}, \textbf{\textcolor{green}{Green}}, and \textbf{\textcolor{blue}{Blue}} boxes correspond to queries that activate most to \textit{Rain}, \textit{Fog} and \textit{Snow} respectively. Best viewed when zoomed in and in color.   }
	\label{explain}
	
\end{figure}

\noindent \textbf{What do the weather queries learn?} The weather queries are embeddings which learn what type of degradation is present in the image. These queries help in predicting the corresponding task vector which is helpful to inject the task information to get a better prediction. To show this, we visualize the attention maps of eight random queries (out of 512) for three images corresponding to different weather degradations in Figure \ref{explain}. It is interesting to observe that queries \textit{Q1}, \textit{Q3}, and \textit{Q6} activate highly for foggy image. They attend throughout the image to all the places afflicted by the fog. Queries \textit{Q2}, \textit{Q4} and \textit{Q8} are observed to activate highly for rainy images and the attention maps are sparse corresponding to the rain details. Similarly, queries \textit{Q5} and \textit{Q7} activate to snow images more when compared to images with rain and fog. This shows that different queries activate for different weather degredations helping TransWeather learn the underlying weather condition and give better predictions. It can also be noted that when an image is degraded by multiple weather conditions, multiple task type queries activate to encode specific tasks. This can be observed from the middle row of Figure \ref{explain} where queries that attend to both fog and rain activate as the image is degraded by both of these conditions. 

\noindent \textbf{Inference Time:} In Figure \ref{teaser} (bottom row), we compare the inference speed in terms of seconds. The time reported in the table corresponds to the time taken by each model feed forward an image of dimensions $256 \times 256$ during the inference stage. We note that our method is faster (with just 0.14 seconds per image) during inference when compared to the previous weather removal methods. TransWeather has 31 M parameters which are less than that of All-in-One Network which has 44 M parameters.


\noindent \textbf{Differences from All-in-One:} As the All-in-One network \cite{li2020all} is the first method to look into using a  single model instance for all weather removal problems, we present clear differences of our method from All-in-One. First, All-in-One is a CNN-based method while TransWeather uses a transformer backbone built specifically for low-level vision tasks with an extra focus on operating on smaller patches. All-in-One uses multiple encoders while TransWeather utilizes a single encoder. All-in-One uses adversarial training and neural architecture search while TransWeather just uses a combination of L1 and perceptual loss making the training more stable. TransWeather also has a faster inference speed, lesser number of parameters while obtaining better quantitative and qualitative performance.

\begin{figure}[t]
	\centering
	\includegraphics[width=1\linewidth, page=1]{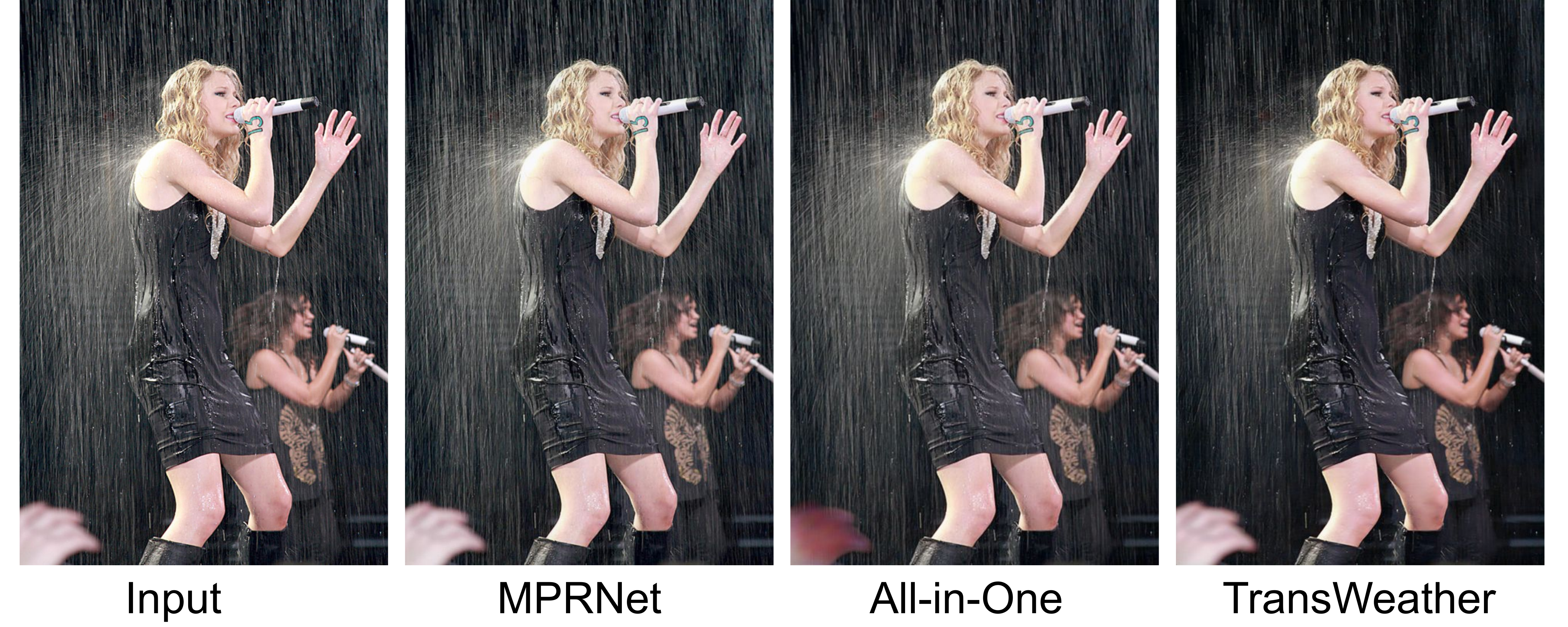} 
	\vskip -10 pt
	\caption{\textbf{Limitations of our method:} High intensity rain and the splattering effect of rain is not removed by any method.  }
	\label{limitation}
	\vspace{-2.3em}
\end{figure}

\noindent \textbf{Limitations:} Although TransWeather achieves better performance than previous methods, there are still some open problems that TransWeather does not solve.  TransWeather does not perform well in some real world images afflicted by high intensity rains. This can be understood as sometimes real-rain is very different in terms of streak size and intensity and are difficult to model. Moreover, if the intensity of rain is high, it creates a splattering effect when it hits the surface of objects or people in the scene. Removing this splattering effect is still a limitation by all methods including TransWeather. A sample limitation is illustrated in Figure  \ref{limitation}.

\vspace{-1em}

\section{Conclusion}
In this work, we proposed TransWeather - an efficient transformer-based solution for the all adverse weather removal problem. We focus on building a \textit{single model instance} which can remove any weather degradation present in the image. We build a single encoder-decoder network for restoration while using learnable weather type queries in the decoder to learn the type of weather degradation and use that information for the weather removal process. We also propose a novel transformer encoder architecture base which work on sub-patches thus aiding transformers to remove small weather degradations more efficiently. We extensively experiment on multiple synthetic and real-world datasets where we use a \textit{single model instance} to get good results while also obtaining a faster inference speed. We also obtain better visual results when tested on real-world adverse weather images.

\noindent \textbf{Acknowledgement:} This work was supported by an ARO grant W911NF-21-1-0135.

{\small
	\bibliographystyle{ieee_fullname}
	\bibliography{egbib}
}

\end{document}